\documentclass[11pt]{article}

\usepackage[final]{acl}

\usepackage[full]{textcomp}
\usepackage{times}
\usepackage{latexsym}

\usepackage[T1]{fontenc}

\usepackage[utf8]{inputenc}

\usepackage{microtype}

\usepackage{inconsolata}

\usepackage{hyperref}
\usepackage{multirow}
\usepackage{url}
\usepackage{tabularx}
\usepackage{array} 
\usepackage{enumitem}
\usepackage{threeparttable}
\usepackage{graphicx} 
\usepackage{amsmath} 
\usepackage{amssymb}  
\usepackage{algorithmic}
\usepackage{algorithm}
\usepackage{caption}
\usepackage{wrapfig}
\usepackage{booktabs}
\usepackage{colortbl}
\usepackage{xcolor}
\usepackage{marvosym}
\usepackage[frozencache,cachedir=.]{minted}
\usepackage{mdframed}
\usepackage{tcolorbox}
\tcbuselibrary{minted,skins, breakable}
\usepackage{xcolor}
\definecolor{mylightblue}{RGB}{100,149,237} 
\definecolor{iccvblue}{rgb}{0.21,0.49,0.74}

\tcbset{
  enhanced, 
  colback=white!200!black, 
  colframe=mylightblue, 
  colbacktitle=mylightblue, 
  title filled, 
  coltitle=white, 
  fonttitle=\bfseries, 
  arc=3mm, 
  outer arc=3mm, 
  boxrule=0.5mm, 
  toprule=0.5mm, 
  bottomrule=0.5mm, 
  titlerule=0.5mm, 
  drop fuzzy shadow, 
  minted options={ 
    breaklines, 
    fontsize=\footnotesize, 
    linenos, 
    numbersep=2mm, 
    framesep=1.5mm, 
  },
}

\setlist[itemize]{topsep=1pt, itemsep=0.5pt}

\definecolor{sparkle}{RGB}{179, 166, 209}
\definecolor{nonum}{RGB}{255, 215, 0}
\definecolor{loc}{RGB}{135, 206, 250}
\definecolor{dist}{RGB}{60, 179, 113}
\definecolor{dir}{RGB}{244, 164, 96}

\title{\raisebox{-0.7em}{\includegraphics[width=0.06\textwidth]{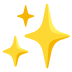}} Sparkle: Mastering Basic Spatial Capabilities in Vision Language Models Elicits Generalization to Spatial Reasoning}

\author{
\textbf{Yihong Tang}\textsuperscript{1$\heartsuit$}, \
\textbf{Ao Qu}\textsuperscript{2\Letter$\heartsuit$}, \
\textbf{Zhaokai Wang}\textsuperscript{3$\heartsuit$}, \
\textbf{Dingyi Zhuang}\textsuperscript{2$\heartsuit$}, \
\textbf{Zhaofeng Wu}\textsuperscript{2} \\
\textbf{Wei Ma}\textsuperscript{4}\textbf{,} \
\textbf{Shenhao Wang}\textsuperscript{5}\textbf{,} \
\textbf{Yunhan Zheng}\textsuperscript{2}\textbf{,} \
\textbf{Zhan Zhao}\textsuperscript{6}\textbf{,} \
\textbf{Jinhua Zhao}\textsuperscript{2} \
\\
[1mm]
\textsuperscript{1}McGill University \quad
\textsuperscript{2}Massachusetts Institute of Technology \quad\\
\textsuperscript{3}Shanghai Jiao Tong University \quad
\textsuperscript{4}The Hong Kong Polytechnic University \\
\textsuperscript{5}University of Florida \quad
\textsuperscript{6}The University of Hong Kong
\\
[0.5mm]
\tt\small yihong.tang@mail.mcgill.ca \quad
\{qua,dingyi\}@mit.edu \quad
\tt\small wangzhaokai@sjtu.edu.cn 
}

\begin{document}

\maketitle

\begin{abstract}
Vision language models (VLMs) perform well on many tasks but often fail at spatial reasoning, which is essential for navigation and interaction with physical environments. Many spatial reasoning tasks depend on fundamental two-dimensional (2D) skills, yet our evaluation shows that state-of-the-art VLMs give implausible or incorrect answers to composite spatial problems, including simple pathfinding tasks that humans solve effortlessly. To address this, we enhance 2D spatial reasoning in VLMs by training them only on basic spatial capabilities. We first disentangle 2D spatial reasoning into three core components: direction comprehension, distance estimation, and localization. We hypothesize that mastering these skills substantially improves performance on complex spatial tasks that require advanced reasoning and combinatorial problem solving, while also generalizing to real-world scenarios. To test this, we introduce \emph{Sparkle}, a framework that generates synthetic data to provide targeted supervision across these three capabilities and yields an instruction dataset for each. Experiments show that VLMs fine-tuned with \emph{Sparkle} improve not only on basic tasks but also on composite and out-of-distribution real-world spatial reasoning tasks. These results indicate that enhancing basic spatial skills through synthetic generalization effectively advances complex spatial reasoning and offers a systematic strategy for boosting the spatial understanding of VLMs.
Source codes of \emph{Sparkle} are available at \url{https://github.com/YihongT/Sparkle}.

\end{abstract}

\begingroup
\renewcommand\thefootnote{}\footnotetext{
$^\heartsuit$ Equal contribution.
\textsuperscript{\Letter}\ Corresponding author.
}
\endgroup

\section{Introduction}

\begin{figure}[t]
  \centering
    \includegraphics[width=\linewidth]{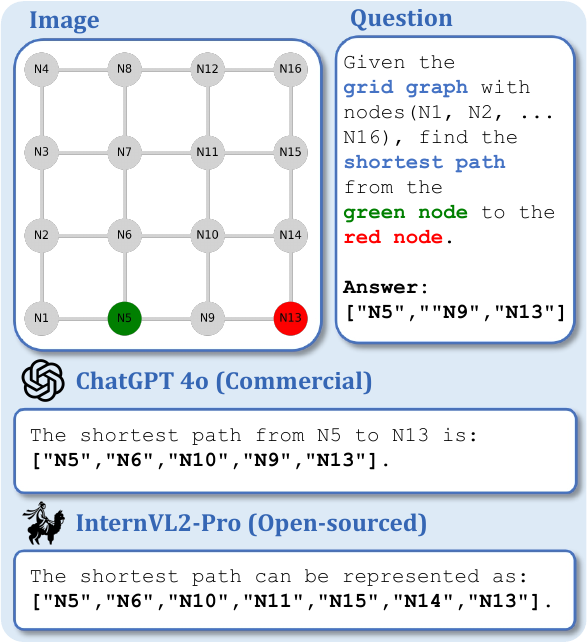}

    \vspace{-1mm}
    \caption{VLMs fail to solve the pathfinding problem
    }
    \label{fig:vlmsb}
\vspace{-4mm}
\end{figure}

\vspace{-1mm}
Vision language models (VLMs)~\cite{gpt4v,VLM:LLaVA,VLM:InternVL-1.5,hong2024cogvlm2,wang2023cogvlm} have demonstrated near-human performance in tasks like image captioning \cite{chen2015cococaption}, visual question answering (VQA) \cite{Datasets:VQAv2,Datasets:TextVQA} and abundant downstream tasks by combining visual and text inputs to reason about the physical world. 
However, these models exhibit significant limitations in understanding spatial relationships. For instance, as shown in Figure~\ref{fig:vlmsb}, state-of-the-art (SoTA) VLMs GPT-4o and InternVL2-Pro~\cite{gpt4v,VLM:InternVL-1.5} generate implausible responses to a shortest path problem that a human could solve at a glance, a simple 2D spatial reasoning task.

Nevertheless, 2D spatial reasoning is essential for VLMs to understand and interact with the physical environments, shaping their ability to solve mazes \cite{ivanitskiy2023configurable,wang2024picture}, plan routes \cite{feng2024citygpt,chen2024mapgpt}, and solve geometric problems like humans \cite{fernandes2009geometric}. 
These tasks emphasize 2D spatial reasoning, requiring VLMs to process and navigate flat visual planes, interpret spatial relationships, and make decisions based on geometric understanding. Such capabilities are fundamental in translating visual input into actionable insights.
While more and more VLMs are developed with larger training datasets and extensive benchmarks \cite{VLM:SEED,zhang2024benchmarking}, the focus on enhancing spatial reasoning has received comparatively less attention, despite its importance to the core capabilities of VLMs. 

In this paper, we study VLMs' spatial reasoning capabilities in a 2D space by investigating three key questions: (1) How well do existing models perform on 2D spatial reasoning? (2) What fundamental tasks affect spatial reasoning capabilities in 2D? (3) Can mastering basic tasks help improve composite and real-world spatial reasoning?

We begin by providing a systematic breakdown of 2D spatial reasoning, grounded in the principles of coordinate systems that represent 2D space. 
From this analysis, we identified three basic capabilities fundamental for spatial reasoning in 2D space: direction comprehension, distance estimation, and localization. A systematic evaluation of the performance of existing open-source and closed-source VLMs on these three basic capabilities reveals that even the most advanced VLMs sometimes struggle with these fundamental tasks. For instance, in a simple 2D direction classification task, where a model is asked to determine the relative direction (top left, top right, bottom left, bottom right) of one object relative to another on a straightforward diagram with only two objects, the state-of-the-art VLM GPT-4o can achieve only 76.5\% accuracy. In contrast, a human should answer these questions correctly with little effort.

Most real-world spatial reasoning tasks, such as pathfinding \cite{lester2005pathfinding,cui2011based}, inherently require the composition of the basic capabilities identified above. A composite task is often subject to specific constraints that necessitate tailored solutions, unlike improving basic spatial reasoning capabilities, which can exhibit generalizability. In order to effectively improve the model's overall spatial reasoning capabilities in 2D space, we raise a conjecture: whether a VLM that masters the three basic capabilities can generalize and perform better on more complex composite spatial tasks. In other words, can a VLM exhibit compositional generalizability~\cite{compositional_generalization} in spatial reasoning tasks? 

To test this, we propose Sparkle, which stands for \textbf{SPA}tial \textbf{R}easoing through \textbf{K}ey capabi\textbf{L}ities \textbf{E}nhancement. This framework fine-tunes VLMs on these three basic spatial capabilities by programmatically generating synthetic data and providing supervision to form an instruction dataset for each capability. 
Additionally, Sparkle creates simplified visual representations to reduce recognition errors, allowing us to focus specifically on enhancing and evaluating VLMs' spatial capabilities.
Our experimental results show that models trained on Sparkle achieve significant performance gains, not only in the basic tasks themselves (e.g., improving from 35\% to 83\% for InternVL2-8B on direction comprehension) but also in generalizing to composite and out-of-distribution general spatial reasoning tasks (e.g., improving from 13.5\% to 40.0\% on the shortest path problem). Additionally, our ablation study confirms the importance of mastering all three basic spatial reasoning capabilities.
To summarize, our contributions are: 
\begin{itemize}[leftmargin=*]
    \item We show that existing VLMs struggle with spatial reasoning tasks that humans solve effortlessly.
    
    \item We identify three basic spatial reasoning components and propose the Sparkle framework to improve these three fundamental spatial reasoning capabilities.

    \item Our experiments prove Sparkle's effectiveness in significantly enhancing the basic spatial capabilities of VLMs, with strong generalizability to out-of-distribution composite and real-world spatial reasoning tasks.
\end{itemize}

\begin{figure*}[t]
    \centering
    \includegraphics[width=\linewidth]{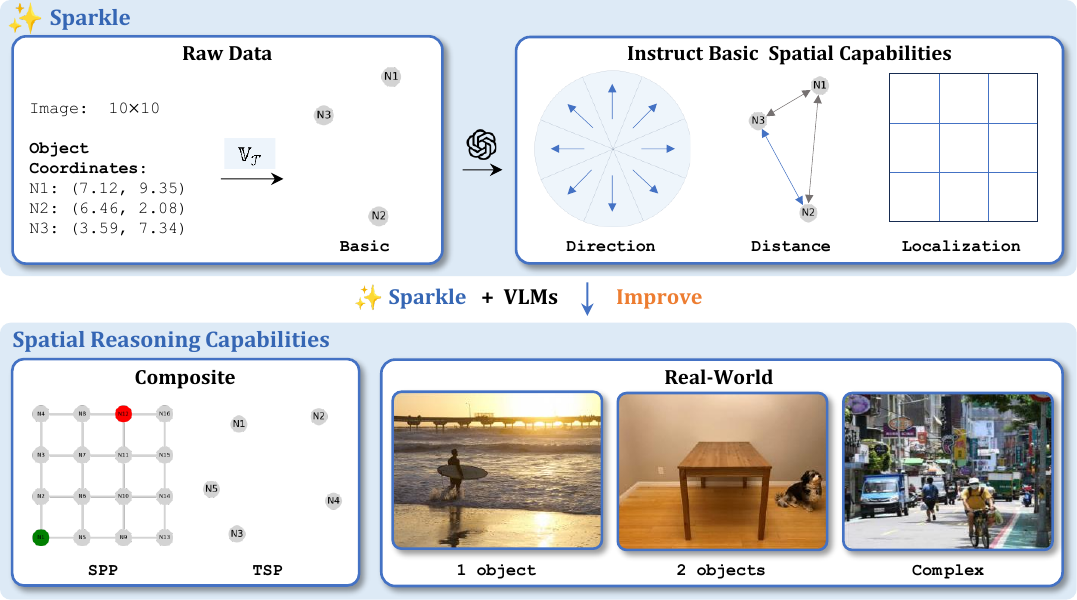}
    \vspace{-4mm}
    \caption{The proposed Sparkle framework.}
    \label{fig:sparkle}
\end{figure*}

\section{Related Work}

\label{sec:related}

\subsection{Vision Language Models and Applications}

Early works on VLMs, such as CLIP~\cite{VLP:CLIP} and ALIGN~\cite{VLP:ALIGN}, leveraged contrastive learning to align visual and textual embeddings in a shared latent space, demonstrating strong capabilities in linking visual content with corresponding natural language descriptions.
With the rapid advancement of Large Language Models (LLMs), modern VLMs increasingly combine pretrained vision models~\cite{TransF:ViT,VLM:InternVL} with powerful LLMs~\cite{TransF:Vicuna,TransF:Qwen,jiang2023think,TransF:InternLM2} to facilitate a more cohesive understanding of both modalities~\cite{VLM:LLaVA,TransF:Qwen,VLM:InternVL-1.5}. 
This approach enables richer visual reasoning, open-ended image captioning, and more interactive multimodal dialogue systems. 

VLMs have been applied in various pre-training tasks, such as image-text matching, masked image modeling, and multimodal reasoning~\cite{VLP:BLIP,VLP:BLIPv2,VLP:BEiTv3}. In downstream tasks, they excel in applications like visual question answering~\cite{antol2015vqa,wang2022ofa,tang2025street}, image captioning~\cite{li2020oscar,yu2024can,sidorov2020textcaps,wang2021confidence}, image generation based on textual prompts~\cite{ramesh2022hierarchical,baldridge2024imagen}, and aiding human-machine interactions in complex real-world settings, showcasing their versatility and potential across a broad range of vision language applications.

\subsection{Spatial Reasoning in LLMs and VLMs}
\label{sec:related_work_spatial_vlms}

Spatial reasoning in LLMs involves understanding and manipulating spatial relationships described in text. Early work focused on extracting spatial information from natural language \cite{hois2011towards,kordjamshidi2011spatial}. Recent efforts emphasize improving multi-hop spatial reasoning \cite{li2024advancing,tang2024itinera}, especially in complex scenarios like 2D visual scenes \cite{shi2022stepgame}. Methods include pretraining on synthetic datasets to better capture spatial patterns \cite{mirzaee2021spartqa}, and using in-context learning to generalize spatial reasoning across tasks, such as transforming spatial data into logical forms or visualizing reasoning trace \cite{yang2023coupling,wu2024visualization}.

Building on these foundations, VLMs extend spatial reasoning by integrating visual inputs and often implicitly encode spatial knowledge through large-scale pretraining on visual-text datasets \cite{VLP:CLIP,li2023scaling}. Early studies on VLMs primarily focus on understanding spatial relationships between objects in front-view images \cite{liu2023visual}, laying the groundwork for 2D spatial reasoning. More recently, research on VLMs has expanded to 3D reasoning tasks, which introduce additional challenges such as depth estimation \cite{chen2024spatialvlm} and path planning \cite{chen2024mapgpt,deng2020evolving}, as seen in applications like robotic grasping \cite{xu2023reasoning} and navigation \cite{shah2023lm,chiang2024mobility} in the embodied AI field \cite{li2024auto}. Despite these advances, 2D spatial reasoning remains more fundamental and flexible, as it can be applied to various tasks, including VQA \cite{VLM:SEED,kamath2023whats,li2024topviewrs} and user interface grounding \cite{rozanova2021grounding}. Due to its broad applicability and foundational role, this work focuses on exploring 2D spatial reasoning capabilities within VLMs.

\begin{figure*}[t]
    \centering
    \includegraphics[width=\linewidth]{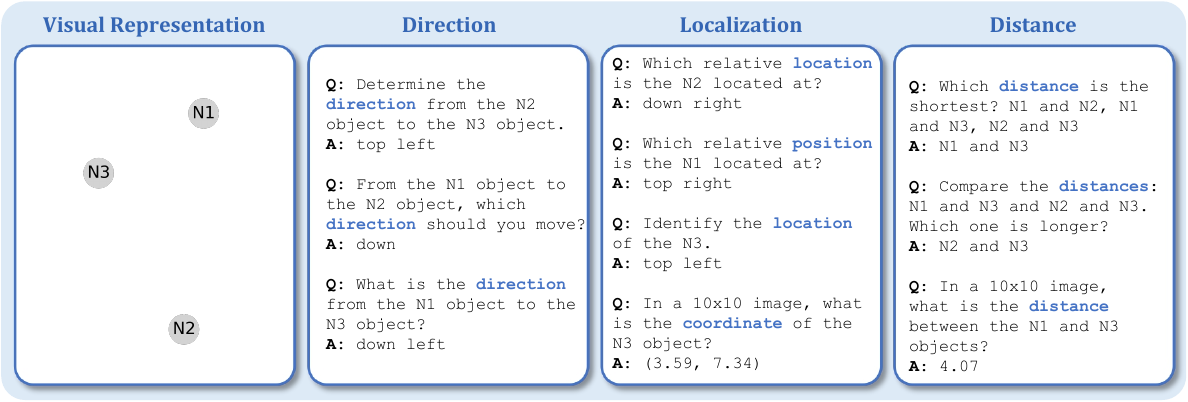}
    \vspace{-5mm}
    \caption{An instruction data sample from Sparkle.}
    \label{fig:sr_instruct_sample}
\end{figure*}

\vspace{-2mm}

\section{Methodology}

In order to systematically evaluate and enhance the spatial reasoning capabilities of VLMs in 2D environments, we introduce the Sparkle framework, as illustrated in Figure~\ref{fig:sparkle}. 

\subsection{Disentangling Spatial Reasoning\label{sec:disentangle}}

2D spaces are usually represented by coordinate systems, which provide a structured way to describe objects' positions and spatial relationships within a plane \cite{byrne1989spatial}. These systems rely on core principles to articulate an object's position: direction defines orientation, distance represents magnitude, and localization integrates both to precisely describe a location \cite{just1985cognitive}. Building on these principles and characteristics of 2D spaces, we identify three foundational components of 2D spatial reasoning: (1) \emph{Direction Comprehension}: The ability to understand the orientation of an object relative to a reference object; (2) \emph{Distance Estimation}: The ability to measure the spatial displacement between objects; (3) \emph{Localization}: The ability to determine the position of an object in space. 
Cognitively, Freksa \cite{freksa1991qualitative} identifies orientation, proximity, and the spatial arrangement of objects as universally useful conceptual properties for spatial reasoning. Frank \cite{frank1992qualitative} also adopted a similar decomposition to study human reasoning about space and spatial properties. The conceptual neighborhoods theory \cite{freksa1991qualitative} demonstrates that simpler conceptual distinctions naturally generalize to broader reasoning contexts (our hypothesis). 
These evidences support the disentangled spatial capabilities form the foundation of 2D spatial reasoning, offering essential elements required to fully describe, understand, and reason about an object's position and relationships with other objects within a 2D space. This decomposition enables a systematic and comprehensive evaluation of spatial reasoning by disentangling these basic spatial capabilities, enhancing specificity in assessing spatial reasoning capabilities in VLMs.

\subsection{Sparkle}

\label{sec:sparkle}


To comprehensively investigate our hypothesis, we introduce Sparkle, a simple yet effective framework for constructing an instruction dataset focused on enhancing a model's spatial reasoning abilities. This framework only improves VLMs' basic spatial capabilities, and this design enables us to evaluate whether models that perform well on basic spatial reasoning tasks can also excel in more complex and composite problems.

\subsubsection{Instruction Data Generation}

The design of our instruction dataset focuses on three basic spatial capabilities: direction, distance, and localization, based on insights provided in Section \ref{sec:disentangle}. The proposed fine-tuning pipeline does not require manual labeling, as all data can be programmatically generated.

We use \(\mathbb{G}\) to denote a data generator that can generate a set of objects, \( P = \{N_i\}_{i=1}^n \), representing a training sample of basic spatial capabilities. Each object \( N_i=(x_i, y_i) \in \mathbb{R}^2 \) consists of randomly sampled coordinates within a bounded region. 
For each basic capability \(\mathcal{T}\in\{\text{dir.}, \text{dist.}, \text{loc.}\}\), we construct a dataset \( D_{\mathcal{T}} \) containing input-output pairs \((\mathcal{X}^{\mathcal{T}}, \mathcal{Y}^{\mathcal{T}})\), where \( \mathcal{X}^{\mathcal{T}} \) represents the inputs and \( \mathcal{Y}^{\mathcal{T}} \) represents the corresponding ground truth outputs. Each input \( \mathcal{X}^{\mathcal{T}} \) consists of: (1) A visual input \( \mathcal{X}_V^{\mathcal{T}} \): A labeled diagram representing the spatial configuration of a sample of objects through a visual representation function \( \mathbb{V}_{\mathcal{T}}(P) \), (2) A language prompt \( \mathcal{X}_L^{\mathcal{T}} \): A question querying some aspects of spatial properties for \( P \).
For example, to craft a training sample for direction comprehension, two objects, \(N_1\) and \(N_2\), are selected from \(P\), and a question such as ``What is the direction of \(N_2\) relative to \(N_1\)?'' is posed. The corresponding correct answer \(Y^{\mathcal{T}}\) can be easily computed since we can access the exact coordinates of these objects, e.g., we can obtain the answer to the above question by calculating the vector from \(N_1\) to \(N_2\) based on their coordinates and map it to the corresponding directional label. 
Details are in Appendix~\S\ref{appx:imple}.

The resulting training dataset consists of these generated questions and answers, paired with the corresponding visual representations, as shown in Figure~\ref{fig:sr_instruct_sample}. Specifically, the training pairs are represented as \( \{(\mathcal{X}_{L}^{\text{train}}, \mathcal{X}_{V}^{\text{train}}, \mathcal{Y}^{\text{train}})\} \), where \(\mathcal{X}_{L}^{\text{train}}\) represents the language-based queries, \(\mathcal{X}_{V}^{\text{train}}\) represents the visual representations, and \(\mathcal{Y}^{\text{train}}\) represents the corresponding answers. 
We provide a complete training sample from Sparkle in Appendix~\S\ref{appx:sample_sparkle}.

\subsubsection{Instruction Finetuning for Basic Tasks} 
To enhance the spatial reasoning capabilities of VLMs, we use the Sparkle training set, denoted as \(\mathcal{X}^{\text{train}} = \{(\mathcal{X}_{L}^{\text{train}}, \mathcal{X}_{V}^{\text{train}})\}\). The objective is to minimize the negative log-likelihood of the predicted answers. The loss function \(\mathcal{L}\) is defined as:
\vspace{-1mm}
\[
\mathcal{L}(\theta) = - \mathbb{E}_{(\mathcal{X}^{\text{train}}, \mathcal{Y}^{\text{train}})} \left[ \log p(\mathcal{Y}^{\text{train}} \mid \mathcal{X}_{V}^{\text{train}}, \mathcal{X}_{L}^{\text{train}}; \theta) \right]
\]
where \(\theta\) represents the parameters of the VLM. The training aims to improve the model's proficiency in basic spatial reasoning tasks, which subsequently allows for evaluation of its performance on more complex spatial challenges.

\begin{figure*}[t]
    \centering
    \includegraphics[width=\linewidth]{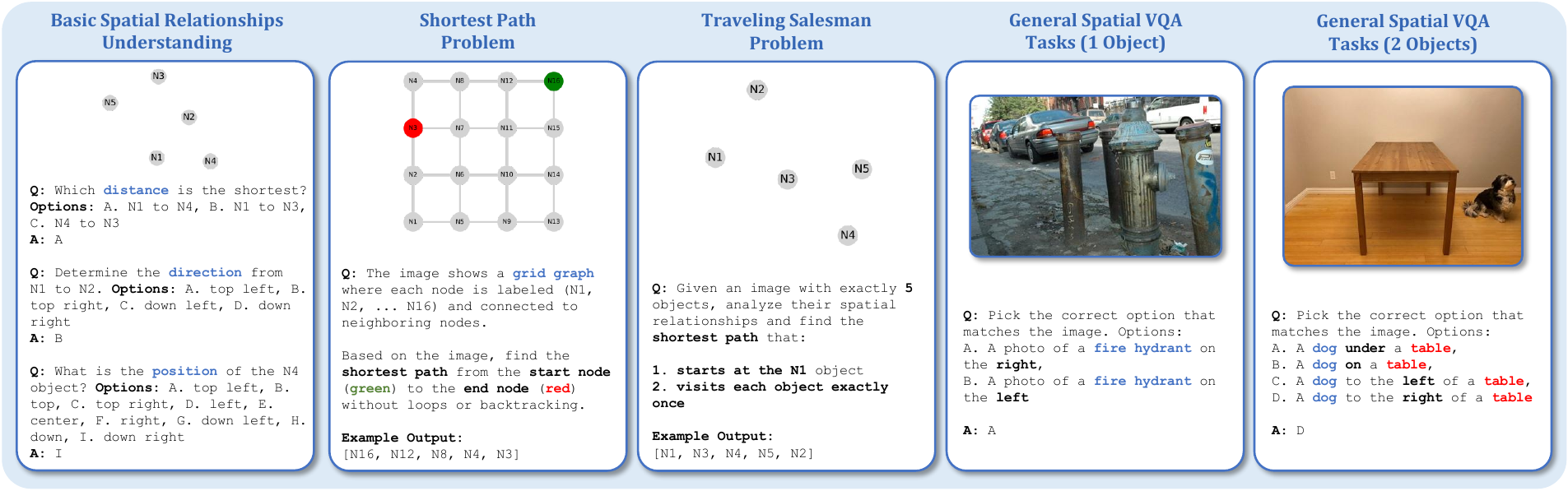}
    \vspace{-5mm}
    \caption{Evaluation samples used in our experiments.}
    \label{fig:sr_eval_sample}
\vspace{-4mm}
\end{figure*}

\subsection{Tasks\label{sec:sparkle_eval}}

The goal of the employed tasks is to evaluate the 2D spatial reasoning capabilities of VLMs and provide a foundation for studying how acquiring basic spatial capabilities can enhance performance on complex tasks. To achieve this, we follow key design criteria: (1) focus on spatial reasoning, and (2) progression from basic to composite tasks.

\subsubsection{Basic Tasks}
As shown in Figure \ref{fig:sr_eval_sample} (left), the basic tasks in Sparkle are designed to assess the model's understanding of three basic spatial capabilities: (1) direction comprehension, (2) distance estimation, (3) localization.
In each basic task, the VLM is provided with an image containing several labeled data objects and a multiple-choice question about the spatial properties of these objects, with the goal of having the model answer these questions correctly. We first generate labeled diagrams that serve as visual inputs, then generate the questions (in multiple-choice format) and corresponding answer pairs to obtain the basic task test set.

\subsubsection{Composite Tasks}

Composite tasks test whether the model can integrate basic spatial skills to solve more complex problems, rather than learning each skill in isolation. 
We use the Shortest Path Problem (SPP) and Traveling Salesman Problem (TSP) for evaluation.


\paragraph{Shortest Path Problem (SPP)}
SPP evaluates the ability to compute the most efficient route between two objects on a 2D grid, requiring a combination of distance estimation and spatial planning.
Consider a grid \( G \) of size \( n \times n \), with two special objects: the start object \( N_{\text{start}} = (x_s, y_s) \) and the end object \( N_{\text{end}} = (x_e, y_e) \). We employ a language model \(\mathrm{LM}\) generates the prompt \( \mathcal{X}_L^{\text{spp}} \) using a predefined prompt template \( \mathbb{P}_{\text{spp}} \), expressed as: \( \mathcal{X}_L^{\text{spp}} = \mathrm{LM}(\mathbb{P}_{\text{spp}}(G, N_{\text{start}}, N_{\text{end}})) \). The visual input is produced similar to basic tasks: \( \mathcal{X}_V^{\text{spp}}=\mathbb{V}_{\text{spp}}(G, N_{\text{start}}, N_{\text{end}}) \).
The combined input for the VLM is \( \mathcal{X}^{\text{spp}} = (\mathcal{X}_V^{\text{spp}}, \mathcal{X}_L^{\text{spp}}) \), and the model is expected to predict the shortest path \(\widehat{\mathcal{Y}}^{\text{spp}}\), which is evaluated against the true shortest path, \(\mathcal{Y}^{\text{spp}}\), computed using standard algorithms.

\paragraph{Traveling Salesman Problem (TSP)}
As shown in Figure \ref{fig:sr_eval_sample} (middle), the TSP represents a more challenging spatial reasoning task, involving combinatorial optimization. The model must find the shortest possible route that visits each object exactly once and returns to the starting object.
Given \( n \) objects \( P^{\text{tsp}} = \{N_i\}_{i=1}^n \) sampled from \(\mathbb{G}\), the ground truth solution \(\mathcal{Y}^{\text{tsp}}\) is computed using a TSP solver \(\mathbb{M}_{\text{tsp}}(P^{\text{tsp}})\). Similarly, the input to VLMs consists of a visual representation \( \mathcal{X}_V^{\text{tsp}} = \mathbb{V}_{\text{tsp}}(P^{\text{tsp}}) \) and a corresponding language prompt \( \mathcal{X}_L^{\text{tsp}} \). The complete input query is \( \mathcal{X}^{\text{tsp}} = (\mathcal{X}_V^{\text{tsp}}, \mathcal{X}_L^{\text{tsp}}) \). Similarly, the model’s predicted order of visiting all objects, \(\widehat{\mathcal{Y}}^{\text{tsp}}\), is then evaluated against the ground truth solution \(\mathcal{Y}^{\text{tsp}}\).

\paragraph{Discussion}
\label{sec:benchmark_discussion}

Given that the SPP can be solved in polynomial time, we expect that if the model can effectively combine its knowledge of basic spatial concepts, it will show significant improvements in solving this task efficiently. On the other hand, the TSP is an NP-hard problem, requiring combinatorial optimization to obtain the exact solution. We include the TSP to push the limits of the model's spatial reasoning capabilities, aiming to investigate how well the model can manage more complex problem-solving tasks beyond the basic integration of spatial skills.

\subsubsection{General Visual-Spatial Tasks}

Sparkle uses simplified visual representations to focus on improving the spatial reasoning abilities of vision-language models (VLMs). The goal is for these enhanced spatial capabilities to generalize across different visual distributions. To evaluate this, we incorporate visual-spatial tasks with real-world images from standard VQA datasets.

\renewcommand{\arraystretch}{1.2}
\begin{table*}[!t]
\centering 
\caption{VLM performance on spatial reasoning tasks before and after Sparkle enhancement. $\Delta$  indicates the relative improvement.} 
\vspace{-2mm}
\resizebox{\textwidth}{!}{

\begin{tabular}{l|ccc|cccc|ccccc}
\toprule
 & \multicolumn{3}{c|}{Basic Tasks} & \multicolumn{4}{c|}{Composite Tasks} & \multicolumn{5}{c}{General Tasks} \\ \cline{2-13}
\multirow{1}{*}{Model} & \multirow{2}{*}{Loc.} & \multirow{2}{*}{Dist.} & \multirow{2}{*}{Dir.} & \multicolumn{2}{c}{SPP} & \multicolumn{2}{c|}{TSP} & \multirow{2}{*}{What's Up} & \multicolumn{2}{c}{COCO-Spatial} & \multicolumn{2}{c}{GQA-Spatial} \\ 
 &  &  &  & 4Grid & 5Grid & 4Obj & 5Obj &  & 1Obj & 2Obj & 1Obj & 2Obj \\ \midrule
GPT-4o & 68.2 & 43.2 & 77.2 & 75.2 & 76.2 & 23.4 & 21.5 & 95.9 & 88.2 & 49.7 & 89.4 & 63.6 \\
Gemini & 61.4 & 41.2 & 56.2 & 66.4 & 64.2 & 14.3 & 16.4 & 69.4 & 50.8 & 34.1 & 42.9 & 21.7 \\
\midrule
LLaVA1.6-7B & 25.2 & 37.3 & 30.8 & 1.7 & 0.9 & 12.0 & 4.0 & 44.9 & 82.3 & 68.5 & 82.7 & 80.4 \\
\ \ \ + Sparkle & 40.7 & 57.2 & 75.9 & 6.2 & 2.3 & 15.8 & 6.8 & 51.4 & 86.8 & 84.2 & 92.3 & 84.1 \\
\ \ \ $\Delta$ & +61.5\%  & +53.4\%  & +146.4\% & +264.7\% & +155.6\% & +31.7\%  & +70.0\%  & +14.5\% & +5.5\%  & +22.9\% & +11.6\% & +4.6\% \\ 
\midrule
Qwen-VL-7B & 25.0 & 37.6 & 24.4 & 2.2 & 1.2 & 11.7 & 3.7 & 42.7 & 89.8 & 74.3 & 98.5 & 94.0 \\
\ \ \ + Sparkle & 59.6 & 61.3 & 64.8 & 5.4 & 4.6 & 18.4 & 12.0 & 49.6 & 96.8 & 87.1 & 99.0 & 96.4 \\
\ \ \ $\Delta$  & +138.4\% & +63.0\%  & +165.6\% & +145.5\% & +283.3\% & +57.3\%  & +224.3\% & +16.2\% & +7.8\%  & +17.2\% & +0.5\%  & +2.6\%  \\ 
\midrule
ChatGLM-4V-8B & 49.7 & 45.7 & 41.6 & 15.8 & 8.7 & 9.8 & 4.4 & 96.4 & 75.9 & 66.5 & 78.3 & 75.5 \\
\ \ \ + Sparkle & 72.6 & 70.3 & 67.9 & 36.3 & 17.1 & 20.4 & 8.6 & 98.4 & 85.4 & 82.9 & 90.5 & 81.8 \\
\ \ \ $\Delta$ & +46.1\%  & +53.8\%  & +63.2\%  & +129.7\% & +96.6\%  & +108.2\% & +95.5\%  & +2.1\%  & +12.5\% & +24.7\% & +15.6\% & +8.3\% \\ 
\midrule
MiniCPM-V2.5-8B & 42.5 & 26.2 & 44.2 & 16.4 & 11.4 & 14.7 & 4.2 & 76.2 & 70.1 & 73.1 & 80.3 & 53.3 \\
\ \ \ + Sparkle & 66.0 & 82.0 & 79.6 & 31.9 & 14.0 & 17.2 & 13.9 & 80.2 & 88.0 & 88.7 & 91.8 & 79.4 \\
\ \ \ $\Delta$ & +55.3\%  & +213.0\% & +80.1\%  & +94.5\%  & +22.8\%  & +17.0\%  & +231.0\% & +5.2\%  & +25.5\% & +21.3\% & +14.3\% & +49.0\% \\ 
\midrule
InternVL2-8B & 61.3 & 44.2 & 34.6 & 15.4 & 13.9 & 17.1 & 9.6 & 92.7 & 92.5 & 71.3 & 97.5 & 85.3 \\
\ \ \ + Sparkle & 74.4 & 83.8 & 83.2 & 38.8 & 39.0 & 21.6 & 14.4 & 94.9 & 94.2 & 78.7 & 99.0 & 90.4 \\
\ \ \ $\Delta$ & +21.4\%  & +89.6\%  & +140.5\% & +151.9\% & +180.6\% & +26.3\%  & +50.0\%  & +2.3\%  & +1.8\%  & +10.4\% & +1.5\%  & +6.0\% \\ \bottomrule
\end{tabular}
}
\label{tab:basic_composite_merged}
\vspace{-4mm}
\end{table*}

\section{Experiments}

In this section, we provide our findings and results to demonstrate the effectiveness of the Sparkle framework. Specifically, the experiments are designed to answer the following research questions:
\textbf{RQ1}: Can mastering basic 2D spatial components enhance overall spatial reasoning capability in VLMs?
\textbf{RQ2}: What insights from the results of evaluations (Section \ref{exp:bench}), enhancements (Section \ref{exp:enhance}), and spatial components (Section \ref{exp:ablation}) can guide improvements in model design, training, and data collection for spatial reasoning in VLMs?

\subsection{Settings}
\label{exp:settings}

\paragraph{Models}

We tested open-source and commercial models to evaluate and enhance VLMs' spatial reasoning capabilities. For commercial VLMs, we used GPT-4o from OpenAI~\cite{VLM:GPT-4v} and Google-Gemini~\cite{VLM:Gemini}. We included LLaVA1.6~\cite{VLM:LLaVA-1.6}, Qwen-VL~\cite{bai2023qwenvl}, ChatGLM-4V~\cite{glm2024chatglm}, MiniCPM-llama3-V2.5~\cite{yao2024minicpm} and InternVL2~\cite{VLM:InternVL-1.5} for open-source models. For all adopted tasks, we report accuracy as the evaluation metric. We use the MS-Swift library \cite{zhao2024swift} and apply the LoRA \cite{TL:LoRA} fine-tuning strategy, with low-rank dimension of 32. We set a constant learning rate of 1e-4 and a batch size of 1. All training and evaluation are performed on GPU clusters with 8$\times$NVIDIA A100 GPUs. 
More details are in Appendix~\S\ref{appx:imple}.


\paragraph{Data}
We built the Sparkle training dataset by generating 10K synthetic images, each paired with spatial reasoning instructions and answers covering three capabilities: direction, distance, and localization, resulting in 170K training samples in total (see detailed statistics and examples in Appendix~\ref{appx:datastat} and Figure~\ref{fig:sample_instruct} in Appendix~\ref{appx:sample_sparkle}). Numerical values were learned using a standard autoregressive cross-entropy loss, standard in visual grounding tasks~\cite{chen2023internvl,liu2023llava}.

We evaluated VLMs on (1) shortest path problem (SPP), (2) traveling salesman problem (TSP), and (3) basic spatial relationship understanding tasks, comprising 2,000 samples each. Additionally, we assessed generalizability by testing performance on out-of-distribution, real-world spatial reasoning benchmarks, including What’s Up~\cite{kamath2023whats}, COCO-spatial~\cite{lin2014microsoft}, and GQA-spatial~\cite{Datasets:GQA}. Further details of task setups and variations are provided in Appendix~\ref{appx:datastat}.

\subsection{Main Results}

\paragraph{Evaluation of Existing VLMs}

\label{exp:bench}

From Table~\ref{tab:basic_composite_merged}, we observe that even the state-of-the-art commercial VLMs cannot obtain satisfactory results on composite tasks like SPP and TSP. Open-source models achieve even worse performance ($\leq$25\% accuracy) on these tasks. 

Specifically, LLaVA performs poorly particularly on SPP compared to TSP, which may be attributed to the grid data structure in SPP being more complex for VLMs to perceive, understand, and generate valid paths grounded on the grid compared to ordering just a few objects in TSP, indicating that these VLMs struggle with visual representations involving intricate spatial structures. Performance on the TSP task worsens as the number of objects increases across most models, highlighting the growing difficulty of spatial reasoning with more objects. However, in SPP, we discover that increasing the grid size has little impact on performance, indicating that a larger grid does not increase the difficulty of reasoning. This result aligns with our initial design principles, where SPP was intended to combine basic spatial understanding with straightforward spatial planning. For general tasks involving real-world images, VLMs still struggle to identify correct spatial relationships and perform spatial reasoning, leaving a significant gap compared to human capability.
To delve into how VLMs behave poorly on spatial reasoning tasks, we further examine their performance on basic spatial relationship understanding, i.e. direction, location and localization comprehension. As shown in Table~\ref{tab:basic_composite_merged}, even the state-of-the-art VLM GPT-4o struggles with basic spatial relationship understanding, achieving only 68.2\%, 43.2\%, and 77.2\% accuracy on the direction, distance, and localization tasks, respectively. These findings explain why VLMs underperform on composite and general tasks, as their weak basic spatial capabilities directly hinder their ability to handle more complex spatial challenges.

\paragraph{Effectiveness of Sparkle}

\label{exp:enhance}

To demonstrate the effectiveness of Sparkle, we present results from fine-tuning all selected open-source VLMs using this method. The results reveal significant improvements in both basic and composite tasks, with generalized improvements to general tasks, indicating that 2D spatial reasoning capabilities can be significantly improved when a model masters the basic components of spatial reasoning. When combining these enhanced spatial abilities with the inherent generalizability of VLMs, the performance gains can be effectively extended to complex spatial reasoning tasks in real-world image domains.
Specifically, Sparkle only contains instructions for basic spatial relationship understanding. However, after fine-tuning with this data, VLMs improved in basic spatial reasoning (around 90\%) and showed significant gains (around 120\%) in composite tasks and general tasks (around 12\%). This justifies that improving these basic spatial reasoning capabilities could effectively enhance VLMs' overall spatial reasoning, enabling them to tackle more complex tasks and comprehend more sophisticated visual representations. This outcome also justifies the rationality of adopting a simplified visual representation, with the hope of helping VLMs acquire inherent spatial reasoning capabilities that can transfer to more complex visual representations.

It is worth noting that the TSP involves more complex spatial reasoning than the SPP. However, VLMs find the SPP more challenging because their outputs must be precisely aligned with the grid. In contrast, solving the TSP only requires determining the optimal order of objects.
When comparing the improvements of VLMs on SPP and TSP, we observe that the gains (around 90\%) on TSP are much smaller than those on the SPP task (150\%). One possible explanation is that the TSP involves more complex optimization challenges, which may not be as easily addressed by simply improving basic spatial reasoning skills, as discussed in Section~\ref{sec:benchmark_discussion}. 
This underscores the need for further research into the optimization capabilities of language models, a topic we hope our findings will inspire.

\paragraph{Generalizability}

\label{exp:general}

In the previous subsection, we have shown that spatial reasoning improvements can generalize from simple tasks to more complex ones. In this section, we evaluate this generalization further by testing spatial reasoning performance in an out-of-distribution visual setting to assess whether these enhanced capabilities extend to broader VLM spatial tasks.
Specifically, we explore whether the enhanced spatial reasoning capabilities transfer to other general VLM spatial tasks. 
As shown in Table~\ref{tab:basic_composite_merged}
, there are consistent gains across general VLM benchmarks related to spatial reasoning. For instance, the COCO-spatial and GQA-spatial benchmarks illustrate that current VLMs often struggle to accurately capture spatial relationships between two objects. With our Sparkle framework, this capability is greatly improved. This generalized improvement demonstrates that Sparkle enhances the inherent spatial reasoning capabilities of VLMs, supporting the effectiveness of using simplified visual representations.
These findings indicate that the Sparkle framework offers a simple yet powerful method for enhancing spatial reasoning capabilities in VLMs. Future VLM research could benefit from incorporating Sparkle’s approach by decomposing spatial tasks into foundational skills and systematically improving them in pretraining and fine-tuning stages, thereby enhancing model performance on complex and general spatial reasoning tasks.

\subsection{Ablation Studies}

\label{exp:ablation}


\paragraph{Impact of Training Components}

\renewcommand{\arraystretch}{1.2}
\begin{table}[ht]
\small
\centering 
\caption{Random perturbation results for InternVL2-8B.}
\vspace{-3mm}

\resizebox{\linewidth}{!}{
\begin{tabular}{lccccc}
    \toprule
         Perturbation    & What's Up   & COCO-1 & COCO-2 & GQA-1 & GQA-2 \\ \hline
Direction    & 85.4 & 90.9  & 62.4  & 96   & 76.8 \\
Distance     & 90.5 & 91.4  & 64.6  & 96.4 & 78.8 \\
Localization & 87.6 & 89.6  & 65.8  & 94.6 & 80.5 \\ 
N/A  & 94.9 & 94.2  & 78.7  & 99.0 & 90.4 \\ \bottomrule
\end{tabular}
}

\label{tab:perturb}

\end{table}

To evaluate the impact of different training components, we first conduct random perturbation (i.e., perturb training labels randomly) to InternVL2-8B on each spatial capabilities to justify the derivation of disentangled spatial capabilities. As shown in Table~\ref{tab:perturb}, the VLM’s performance degrades drastically after perturbation, confirming the critical role of each identified basic spatial component.

\begin{figure}[ht]
  \centering
    \includegraphics[width=\linewidth]{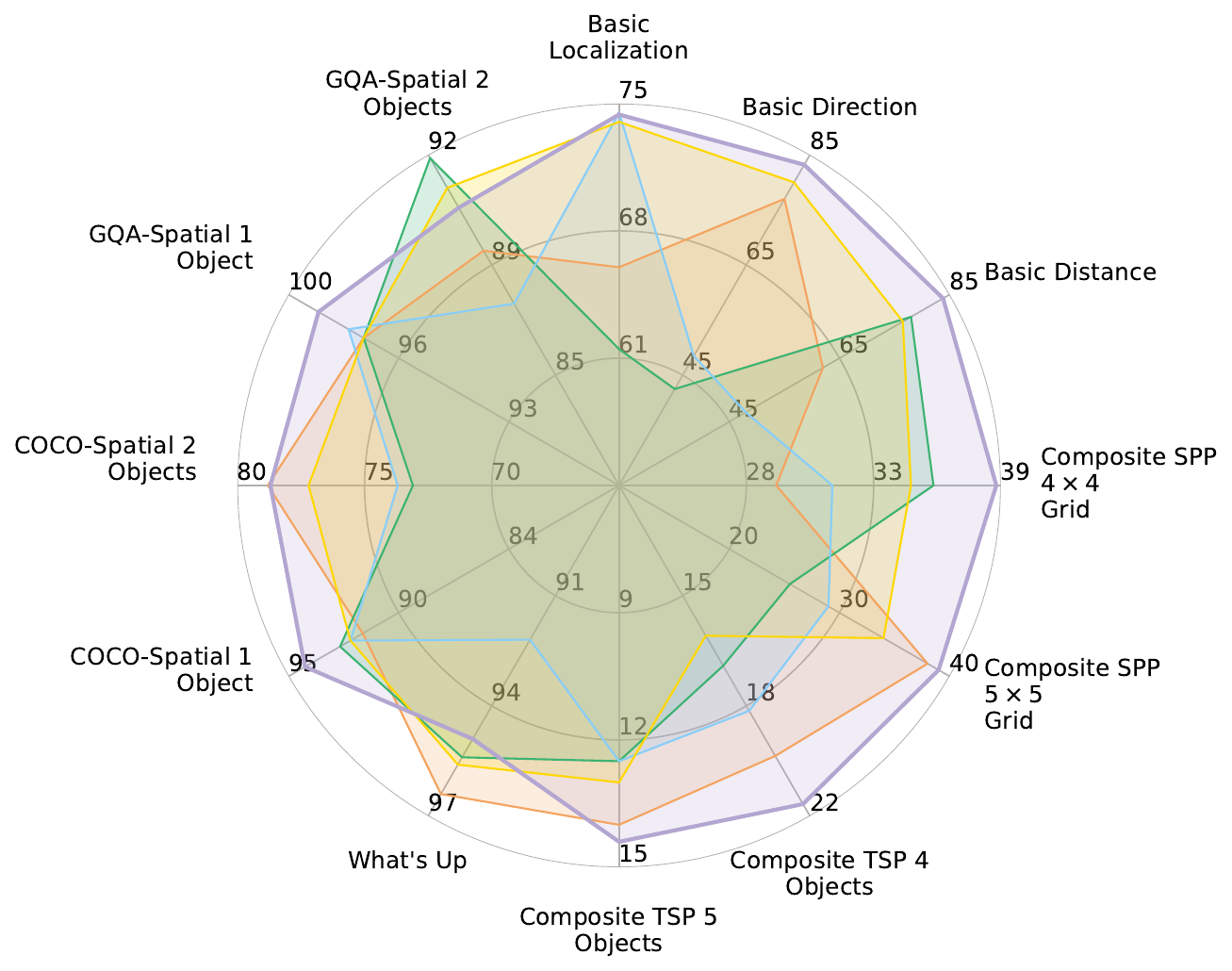}
    \vspace{-6mm}
    \caption{Sparkle variants: Sparkle \textcolor{sparkle}{$\blacksquare$}; Sparkle without numerical information \textcolor{nonum}{$\blacksquare$}; Sparkle (Localization) \textcolor{loc}{$\blacksquare$}; Sparkle (Distance) \textcolor{dist}{$\blacksquare$}; Sparkle (Direction) \textcolor{dir}{$\blacksquare$}. }
    \label{fig:exp_ablation}
\end{figure}

We also trained InternVL2-8B on individual spatial reasoning tasks with our Sparkle framework, resulting in \emph{Sparkle(Direction, Distance, Localization)}. We also tested a version called \emph{Sparkle w/o Num} that excludes numerical information (i.e., distance and location estimation) in Sparkle. All the four variants are trained with the same number of total samples as the full Sparkle model.
The results shown in Figure~\ref{fig:exp_ablation} reveal two key insights:
First, \emph{Sparkle w/o Num} consistently underperforms compared to the full Sparkle model, particularly in tasks that require strong distance reasoning, such as TSP. This suggests that incorporating numerical information during training significantly enhances the model's capability in tasks involving distance reasoning and other related composite challenges. 
Second, training on specific spatial reasoning subsets can sometimes yield optimal performance for certain tasks. For example, \emph{Sparkle (Direction)} achieves 96.4\% accuracy on the What's Up benchmark, indicating that task-specific training can be highly effective. This highlights the importance of tailoring the training process to the unique characteristics of individual tasks. When a task emphasizes a particular spatial reasoning capability, focusing the training data on that aspect can improve performance on the targeted task.
The full Sparkle framework consistently delivers the best results across most benchmarks, demonstrating the effectiveness of a more comprehensive approach to training.

\paragraph{Impact of Training Sample Size}

\begin{figure}[h]
  \centering
    \includegraphics[width=\linewidth]{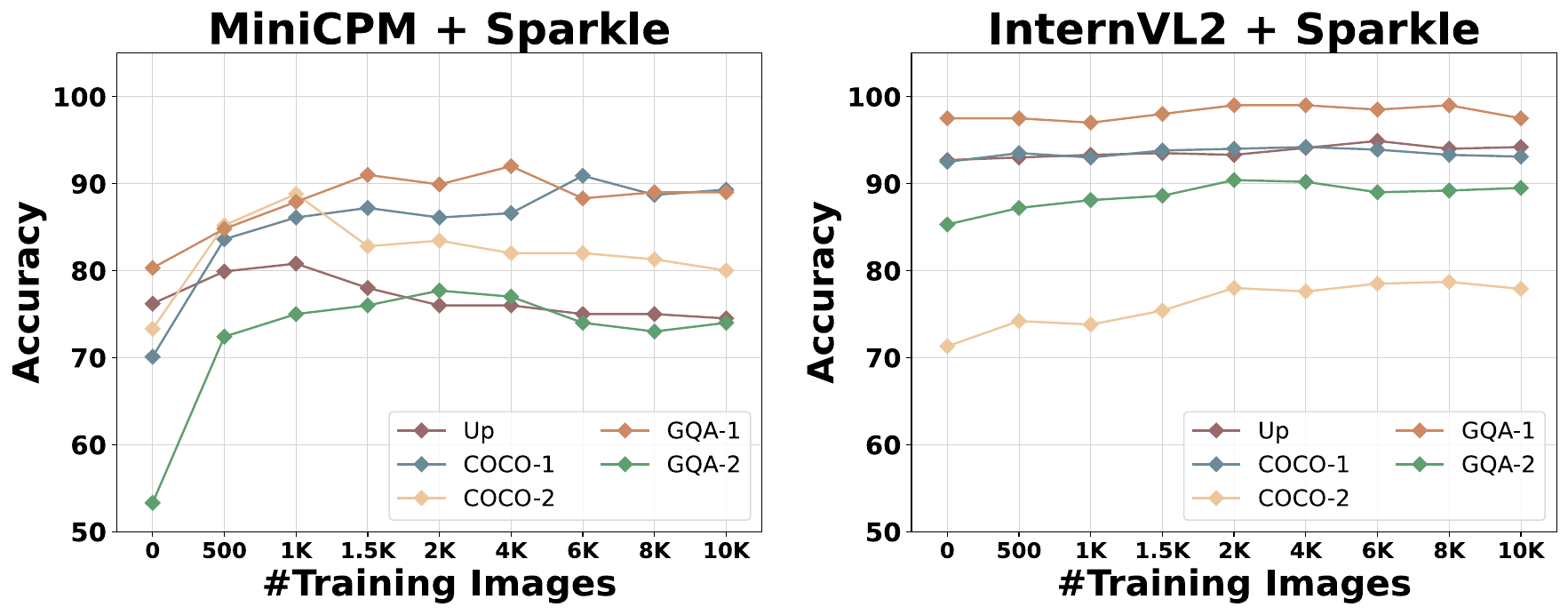}
    \vspace{-6mm}
    \caption{Results of Sparkle on InternVL2 and MiniCPM with varying training sample sizes.}
    \label{fig:exp_trainum}
\end{figure}

We varied the training sample size in Sparkle and evaluated its impact on general spatial reasoning tasks. 
The results are shown in Figure~\ref{fig:exp_trainum}. We observe a general improvement in VLM performance as the training sample size increases, despite some fluctuations in the curve. However, a noteworthy finding is the existence of task-specific sweet spots, beyond which performance gains taper off or degrade. 

\subsubsection{Performances on Common VLM Benchmarks}

\renewcommand{\arraystretch}{1.2}
\begin{table}[ht]
\small
\centering 
\caption{Performance on common VLM benchmarks.}
\vspace{-3mm}

\resizebox{\linewidth}{!}{
\begin{tabular}{l|cc|cc|cc|cc}
    \toprule
    \multirow{2}{*}{Model} & \multicolumn{2}{c|}{SEED-I} & \multicolumn{2}{c|}{MME} & \multicolumn{2}{c|}{BLINK} & \multicolumn{2}{c}{MMBench} \\
    & All & SR & All & Pos & All & SR & All & SR \\
    \hline
    InternVL2 & 75.4 & 62.1 & 1641 & 143 & 50.3 & 80.4 & 82.4 & 46.7 \\
    \ \ +Sparkle & 75.5 & 64.5 & 1644 & 151 & 52.3 & 81.1 & 83.2 & 53.3 \\
    \bottomrule
    \end{tabular}
}

\label{tab:vlmbench}

\end{table}

While VLMs show significant improvements in spatial reasoning with Sparkle enhancement, we also evaluate them on common benchmarks. As shown in Table~\ref{tab:vlmbench}, Sparkle-trained models show substantial improvement in spatially related subdimensions, while maintaining or improving overall performance, demonstrating that Sparkle does not negatively affect the overall abilities of VLMs. This suggests that incorporating Sparkle into the pretraining process could further enhance these general capabilities.

\subsection{Discussion}
\vspace{-1mm}
The results confirm that mastering basic 2D spatial reasoning capabilities through Sparkle can significantly enhance VLMs' overall spatial reasoning in composite tasks (e.g., spatial planning) and general spatial tasks. This directly addresses RQ1 and supports the assumption presented in the methods section. 
Turning to RQ2, the evaluation results revealed the limitations of existing VLMs, particularly in their capability to perceive complex spatial structures, as evidenced in tasks like SPP. This highlights the need for improved model and training designs to support more detailed spatial reasoning.
Moreover, introducing synthetic data focusing on basic spatial relationships has proven to enhance overall VLM spatial reasoning performance, offering a clear path for future spatial data collection. 
Lastly, our ablation study suggests that training specific spatial reasoning capabilities in isolation yields the best results for tasks that demand focused spatial abilities. Therefore, in terms of training strategy, our findings suggest adopting a pre-train and fine-tune approach (i.e., using diverse spatial data in pretraining and fine-tuning specific spatial capabilities tailored to particular tasks) to improve VLMs' performances on corresponding tasks.

\vspace{-1mm}
\section{Conclusion}
\vspace{-2mm}
We present the Sparkle framework to address the limited spatial reasoning ability of Vision Language Models (VLMs). It is designed to enhance spatial reasoning by focusing on three basic capabilities: direction, distance and localization. Experiments show that fine-tuning on these basic capabilities leads to substantial improvements in the basic tasks and composite tasks, showcasing its compositional generalizability. It also leads to generalization on broader tasks, strengthening VLMs’ ability to interact with the physical world.



\section*{Limitations}

While Sparkle shows strong improvements in 2D spatial reasoning, there are still areas for further exploration. Our framework is based on synthetic 2D data with simplified visuals. Although it demonstrates strong performance in 2D spatial problem-solving and generalizes to real-world domains, it may not fully capture the diversity and complexity of real-world imagery. Additionally, the current focus is on basic spatial capabilities. Extending the approach to more complex reasoning, including temporal and 3D spatial understanding, and developing synthetic generalization strategies specifically for 3D spatial tasks remains an open direction. Finally, although we observe promising generalization, further evaluation across broader tasks and domains would strengthen our findings.

\clearpage

\bibliography{main}

\begin{onecolumn}
{
\Large
\textbf{Appendix}\\
\vspace{1.0em}
}
\appendix

\section{Implementation Details}
\label{appx:imple}

We built the Sparkle training dataset by generating 10K images, each with 17 instruction–answer pairs that describe the spatial relationships between objects, resulting in a total of 170K samples. Among these pairs, 3 focus on directions between objects, 7 on distances (including 4 for comparing distances and 3 for estimating numerical distances), and 6 on localization (with 3 for identifying object locations and 3 for estimating exact positions). The final pair describes the overall spatial relationships in the image in natural language. This setup ensures that the VLM maintains its ability to follow instructions effectively. Numerical values are learned using a standard autoregressive cross-entropy loss, as is standard for grounding tasks in VLM training~\cite{chen2023internvl,liu2023llava}. 
A complete sample can be found in Figure~\ref{fig:sample_instruct} in Appendix~\ref{appx:sample_sparkle}.
Our evaluation includes tasks of: (1) shortest path problem (SPP), (2) traveling salesman problem (TSP), and (3) basic spatial relationship understanding. For each of them, we generated 2000 samples, which together make up the evaluation set. For SPP and TSP,  we use LLaMA 3.1~\cite{dubey2024llama} to process the VLMs' responses into list formats to enable metric computation. For the basic tasks, we structured them in a multiple-choice question format.
In addition, for SPP and TSP, we designed experiments that vary by grid size and the number of objects involved. 
Detailed data statistics and sample data are provided in Appendix~\ref{appx:datastat}.
To further assess the generalizability of the improved spatial reasoning capabilities, we evaluated VLMs on existing general spatial reasoning-related benchmarks to examine their out-of-distribution performance. 
We use general benchmarks include What's Up~\cite{kamath2023whats}, COCO-spatial~\cite{lin2014microsoft}, and GQA-spatial~\cite{Datasets:GQA}, featuring real-world images and spatial reasoning questions.

In addition to the experimental settings outlined in Section 4.1, we provide the following categorized implementation details for this work.

For model specifications, the GPT-4o model used in our experiments and demonstrations is based on the gpt-4o-2024-05-13 version, while the Gemini model is Gemini 1.5 Flash. For TSP data generation, we used an open-source Python TSP solver\footnote{https://github.com/fillipe-gsm/python-tsp} to obtain the ground truth visiting order of the given object coordinates.

For VLM evaluations, we focused on four directional categories (top left, top right, bottom left, and bottom right) to make it easier for VLMs to distinguish between directions. 
To discretize object locations for localization learning in VLM, the 2D space is proportionally divided using 40\% and 60\% thresholds along both the \(x\) and \(y\) axes, creating nine distinct regions (center, top, bottom, left, right, top-left, top-right, bottom-left, bottom-right).
Detailed data statistics and distribution visualizations are provided in Section~\ref{appx:datastat}. 

To extract and format the VLMs' responses, we used the LLaMA 3.1 language model~\citep{dubey2024llama},
which converts the results into the required format for metric calculations. The specific prompts used for each task are detailed in Section~\ref{appx:prompt_extract}.
The evaluation for basic spatial relationship understanding is intuitive, as it follows a multiple-choice question format. For the SPP evaluation, we check two criteria: (1) whether the solved path is valid on the grid, and (2) whether the length of the solved path is indeed the shortest between the given start and end objects. For the TSP evaluation, a path is considered ``correct'' only if it exactly matches the solution from the TSP solver mentioned above. To reduce the difficulty for VLMs in solving TSP, we explicitly specify the starting object in our implementation.

For the benchmark evaluation of Vision-Language Models (VLMs), we utilized the following benchmark datasets: MMBench \emph{dev}~\cite{Datasets:MMBench},
SeedBench~\cite{Datasets:Seed-bench}, MME~\cite{Datasets:MME}, and BLINK~\cite{fu2024blink}
Additionally, we employed the VLMEvalKit~\cite{duan2024vlmevalkit}, an open-source evaluation toolkit, to ensure standardized and reproducible evaluation of the VLMs.

\section{Data Statistics}
\label{appx:datastat}

\begin{figure}[h]
    \centering
    \includegraphics[width=0.6\linewidth]{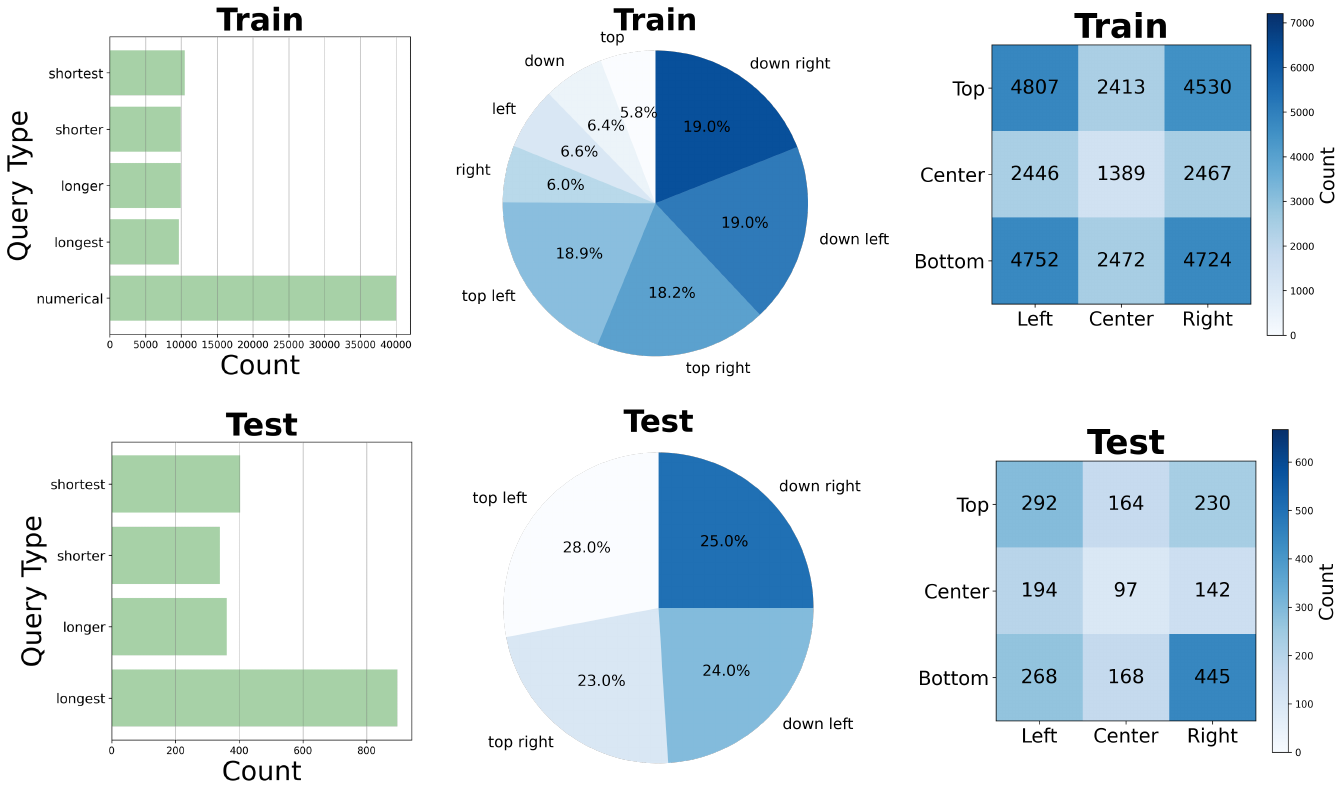}
    \caption{Data statistics of basic spatial relationships (from left to right: distance, direction, and localization statistics).}
    \label{fig:stat_basic}
\end{figure}

To complement Section 4.1, this section provides detailed statistics of the data from Sparkle training set and evaluation. We begin by discussing data related to basic spatial relationships (i.e., distance, direction, localization), covering both Sparkle training set and the spatial relationship understanding task in the evaluation set.

Figure~\ref{fig:stat_basic} illustrates various statistics. In the left column, we see the distribution of questions and instructions related to the \emph{distance} between objects, which includes comparative expressions (e.g., shortest, shorter, longer, longest) and numerical distance estimations considered only in Sparkle training set. The training set shows a fairly even distribution of comparison queries, while in the test set, queries involving the ``shortest'' and ``longest'' distances occur more frequently than those involving ``shorter'' and ``longer''.

The middle column of Figure~\ref{fig:stat_basic} presents the data concerning \emph{directional} relationships between objects. We divided the 2D space into direction sectors: four sectors for testing and eight for training. The directional relationships of ``bottom-right'', ``bottom-left'', ``top-right'', and ``top-left'' each make up about 19\% of the training data, while ``top'', ``bottom'', ``left'', and ``right'' each account for roughly 6\%. In the test set, the four main directional relationships are distributed evenly.

Lastly, the right column in Figure~\ref{fig:stat_basic} shows the \emph{localization} data. Objects are most frequently located in the corners of the space (i.e., top-left, top-right, bottom-left, and bottom-right) in both the training and test sets. The number of objects placed in ``top'', ``bottom'', ``left'', and ``right'' positions is about half that of those in the corners, while the fewest objects are placed in the center. This is due to the intentional narrowing of the center area as we explained in Section~\ref{appx:imple}, which reduces the likelihood of randomly generated objects being placed there. Since there is no clear distinction between regions like ``left'' and ``top-left'', this narrowed design encourages VLMs to accurately distinguish specific areas such as the ``center'', ``top'', ``bottom'', ``left'', and ``right''.

\begin{figure}[h]
    \centering
    \includegraphics[width=0.6\linewidth]{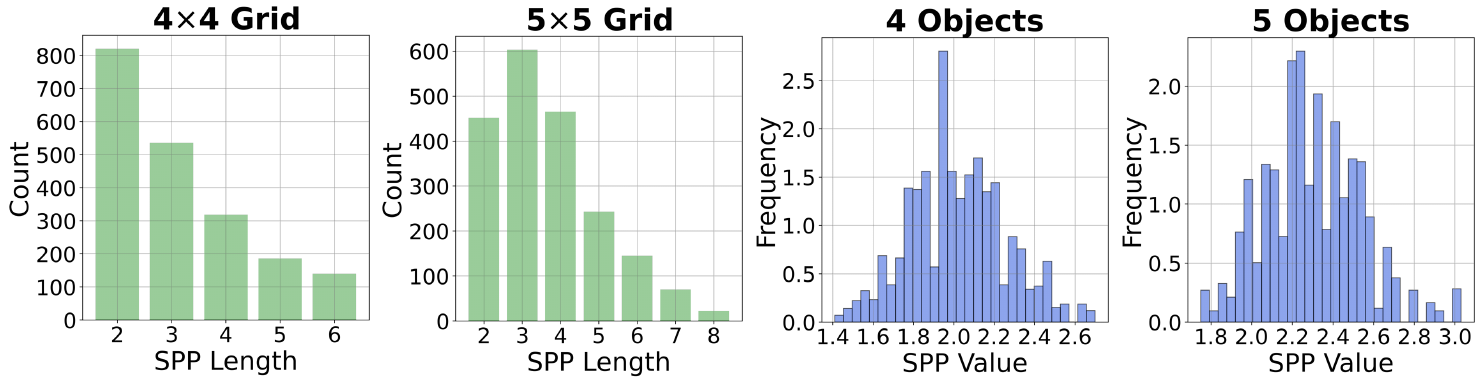}
    \caption{Data statistics of composite spatial reasoning tasks in the evaluation set.}
    \label{fig:stat_composite}
\end{figure}

Figure~\ref{fig:stat_composite} presents data statistics for composite spatial reasoning tasks. The two left subfigures show the distribution of ground truth shortest path lengths in \(4\times4\) and \(5\times5\) grids, while the two right subfigures depict the distribution of total distances for the optimal path in the TSP with 4 and 5 objects.

\clearpage








\section{Prompts for Extracting Inference Results from VLMs}
\label{appx:prompt_extract}

In this section, we provide the designed prompts for a language model to extract results from VLMs' responses. 

\subsection{Multi-choice Questions}

\label{appx:prompt_mcq}

\begin{tcolorbox}[title=Prompt for Extracting Results from VLMs' Responses to Multiple-Choice Questions, breakable]
\begin{minted}{markdown}
Extract the option capital letter from the result and return it as \\boxed{{X}}, where X is the letter. 

Provide no additional content. The result is: ```{result}```. 

Make sure your response is in the \\boxed{{X}} format.
\end{minted}
\end{tcolorbox}

The above prompt is adopted for all evaluations that in a Multi-choice Questions format.

\subsection{Shortest Path Problem}

\label{appx:prompt_spp}
\begin{tcolorbox}[title=Prompt for Extracting Results from VLMs' Responses to Shortest Path Problems, breakable]
\begin{minted}{markdown}
Extract the sequence of node labels from the given input and return it as a Python list.

**Return Format:**
- Do not include any additional text or explanations.
- Ensure that the response is a single list containing only the node text labels (N1, N2, ...).
- If no valid action sequence is found, return 'None'.

**Example Output format:**
```
[node1 text label, node2 text label, ...]
```

Now, extract the result from the following input: ```{result}```. Strictly adhere to the return format.
\end{minted}
\end{tcolorbox}

\subsection{Traveling Salesman Problem}

\label{appx:prompt_tsp}

\begin{tcolorbox}[title=Prompt for Extracting Results from VLMs' Responses to Traveling Salesman Problems, breakable]
\begin{minted}{markdown}
Extract the sequence of movements from the given input and return it as a Python list of object names.

**Return Format:**
- Do not include any additional text or explanations.
- Ensure that the response is a single list containing only the object names.

**Expected Output Format:**
```
{output_format}
```

Now, extract the result from the following response: ```{result}```. Strictly adhere to the output format.
\end{minted}
\end{tcolorbox}

\clearpage

\section{Sample Data Demonstration}

In this part, we provide detailed data sample from our experiments.

\subsection{Data Sample from Sparkle Training Set and Evaluation}

\label{appx:sample_sparkle}

\begin{figure}[h]
  \centering
    \includegraphics[width=\textwidth]{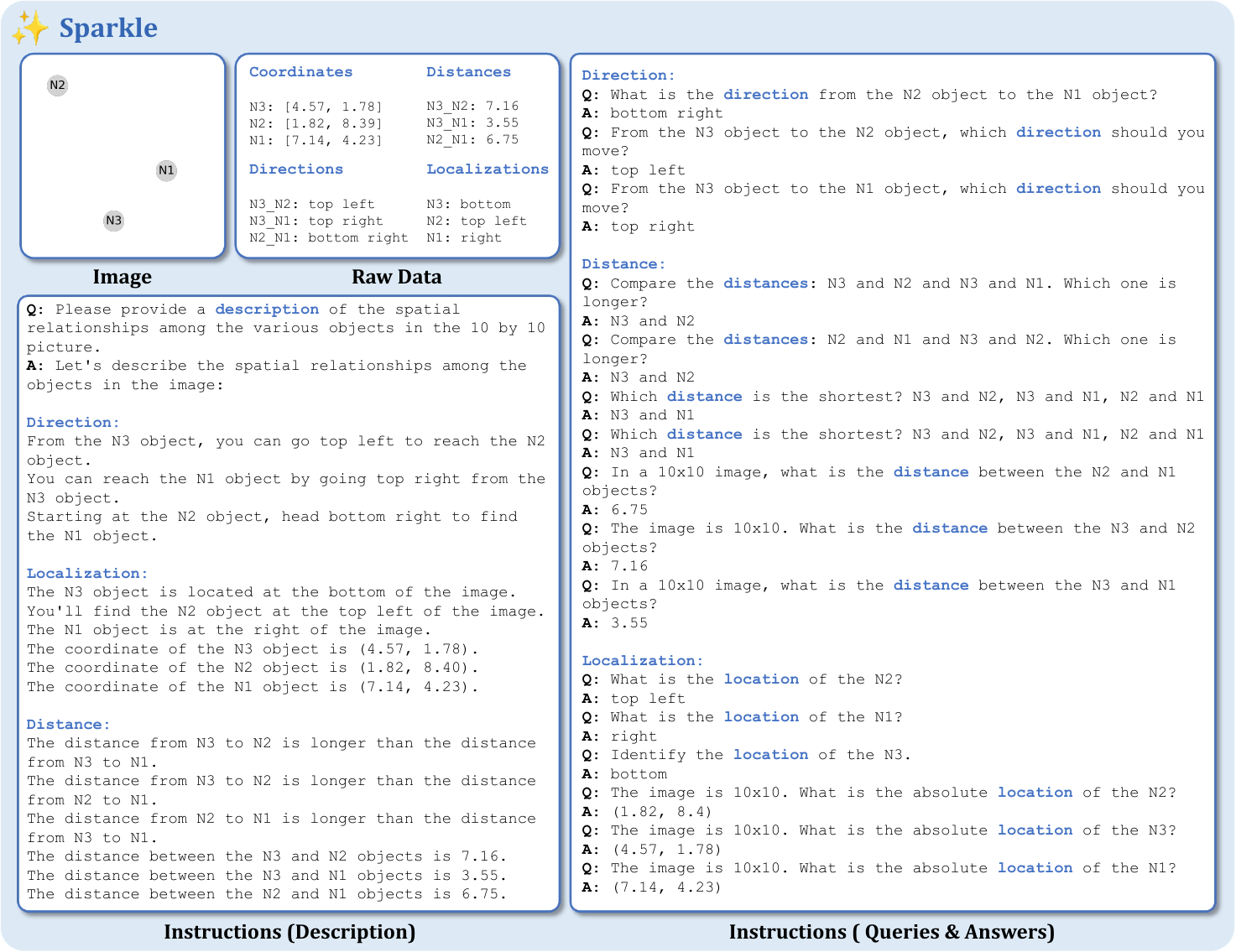}
    \caption{A data sample from the Sparkle training set.}
    \label{fig:sample_instruct}
\end{figure}

\subsection{Data Sample from the Basic Spatial Relationship Understanding task}

\label{appx:sample_basic}

\begin{figure}[h]
  \centering
    \includegraphics[width=\textwidth]{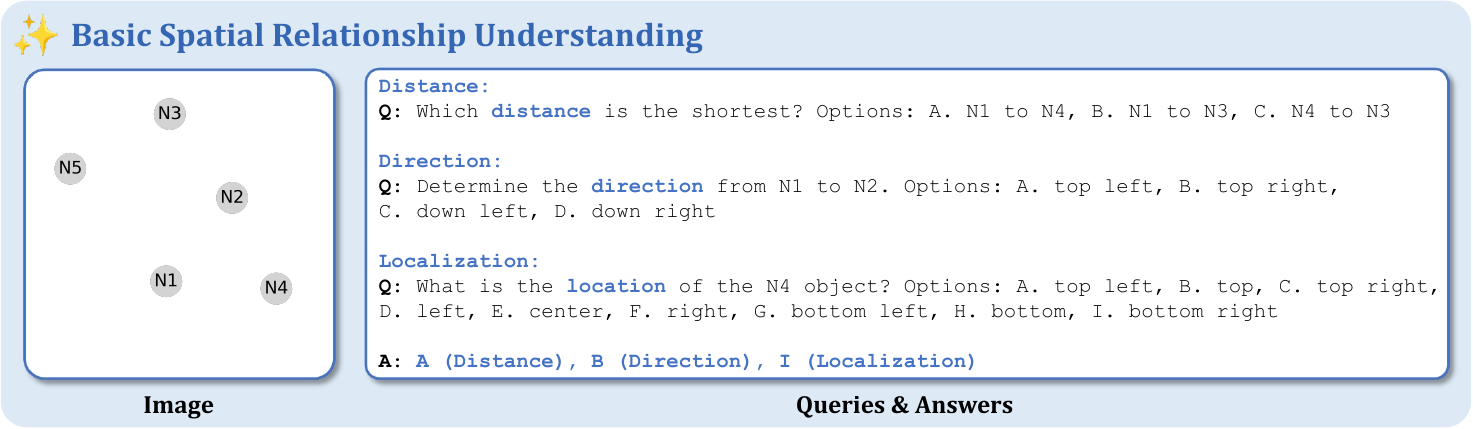}
    \caption{A data sample for Basic Spatial Relationship Understanding}
    \label{fig:sample_instruct}
\end{figure}

\clearpage

\subsection{Data Sample from the Shortest Path Problem}

\label{appx:sample_spp}

\begin{figure}[h]
  \centering
    \includegraphics[width=.75\textwidth]{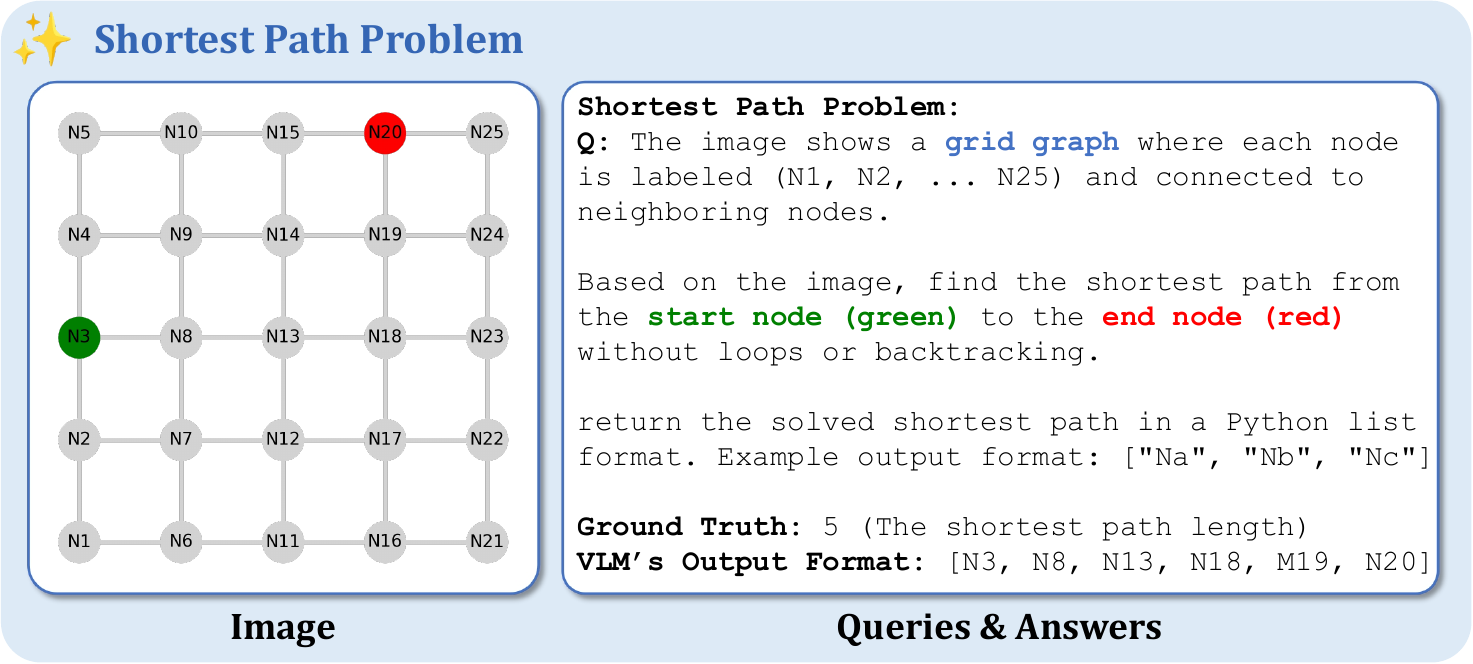}
    \caption{A data sample from the Shortest Path Problem.}
    \label{fig:sample_spp}
\end{figure}

\subsection{Data Sample from the Traveling Salesman Problem}

\label{appx:sample_tsp}

\begin{figure}[h]
  \centering
    \includegraphics[width=.75\textwidth]{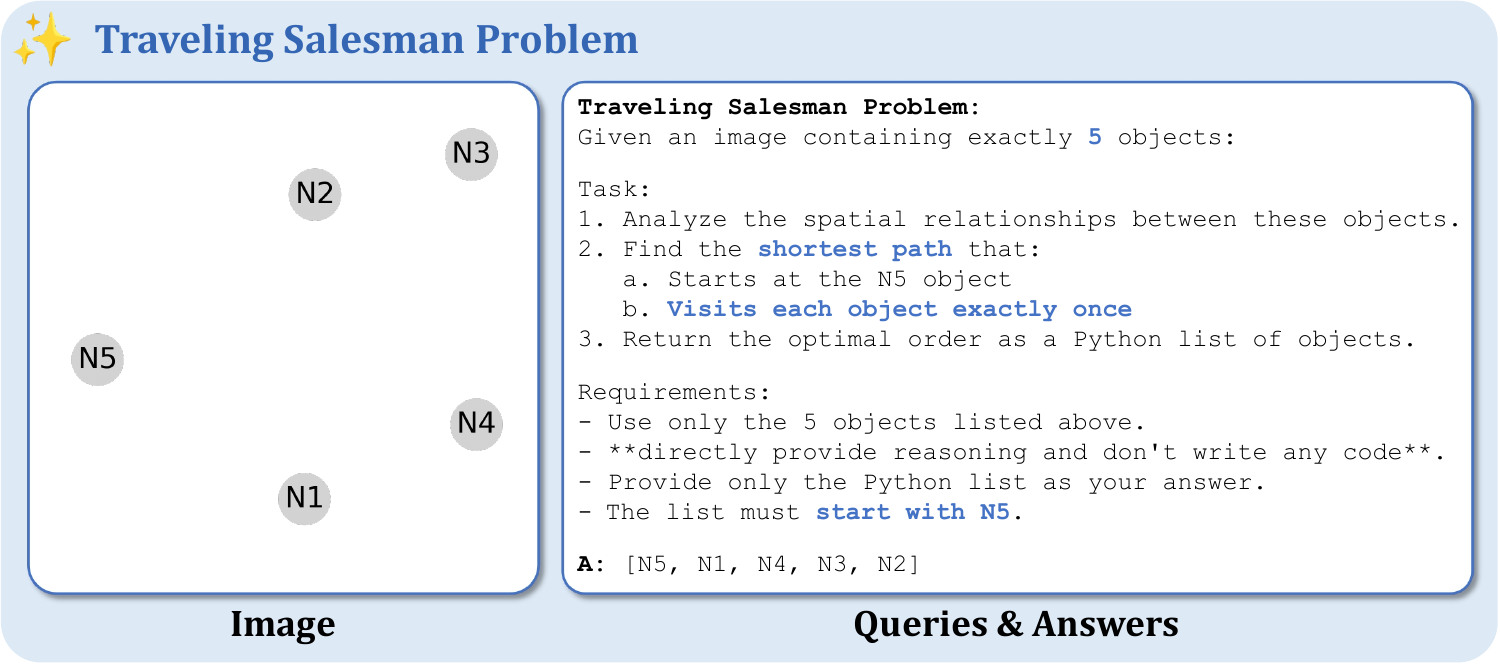}
    \caption{A data sample from the Traveling Salesman Problem.}
    \label{fig:sample_tsp}
\end{figure}

\subsection{Data Sample from General Spatial Tasks}

\label{appx:sample_general}

\begin{figure}[h]
  \centering
    \includegraphics[width=.75\textwidth]{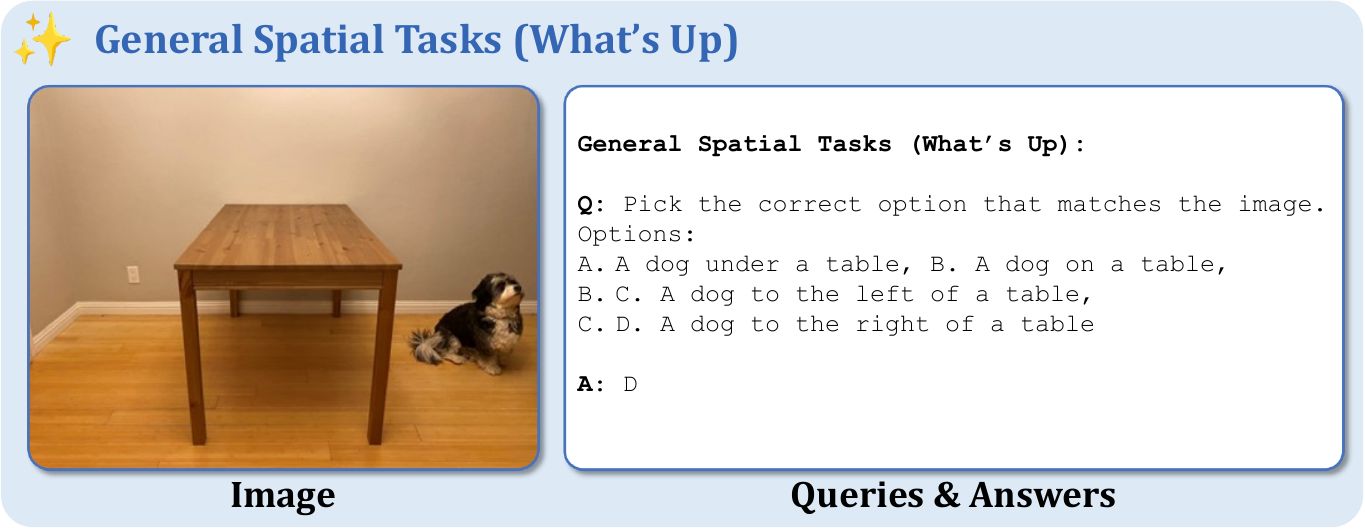}
    \caption{A data sample from the General Spatial Tasks ("What's Up").}
    \label{fig:sample_general}
\end{figure}

\clearpage

\section{Screenshots of Chat with VLMs}

\subsection{GPT-4o}

\begin{figure}[h]
  \centering
    \includegraphics[width=.75\textwidth]{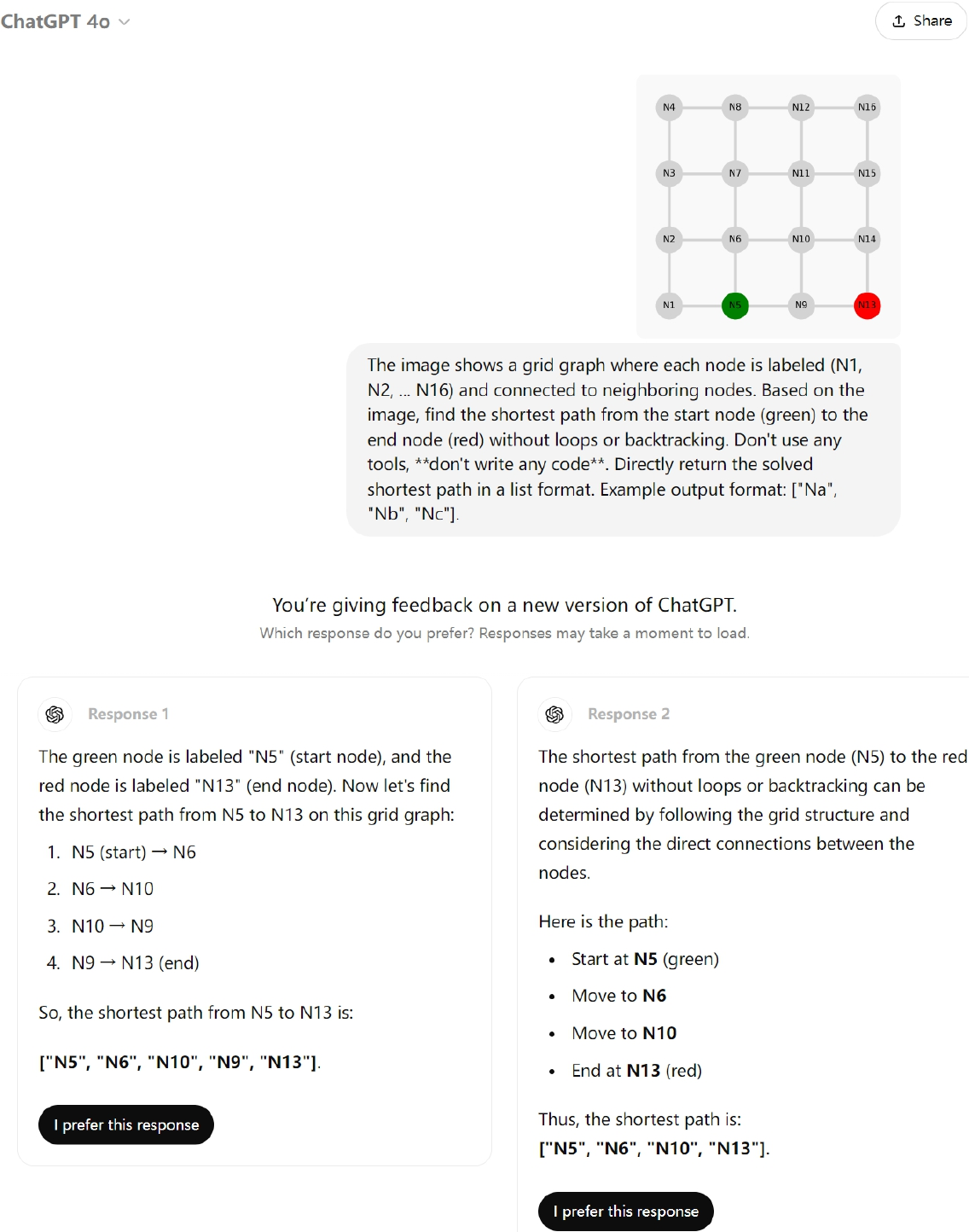}
    \caption{Screenshot supporting Figure 1 in the main paper: Chat interactions with GPT-4o.}
    \label{fig:gpt4o_2}
\end{figure}

\begin{figure}[h]
  \centering
    \includegraphics[width=.75\textwidth]{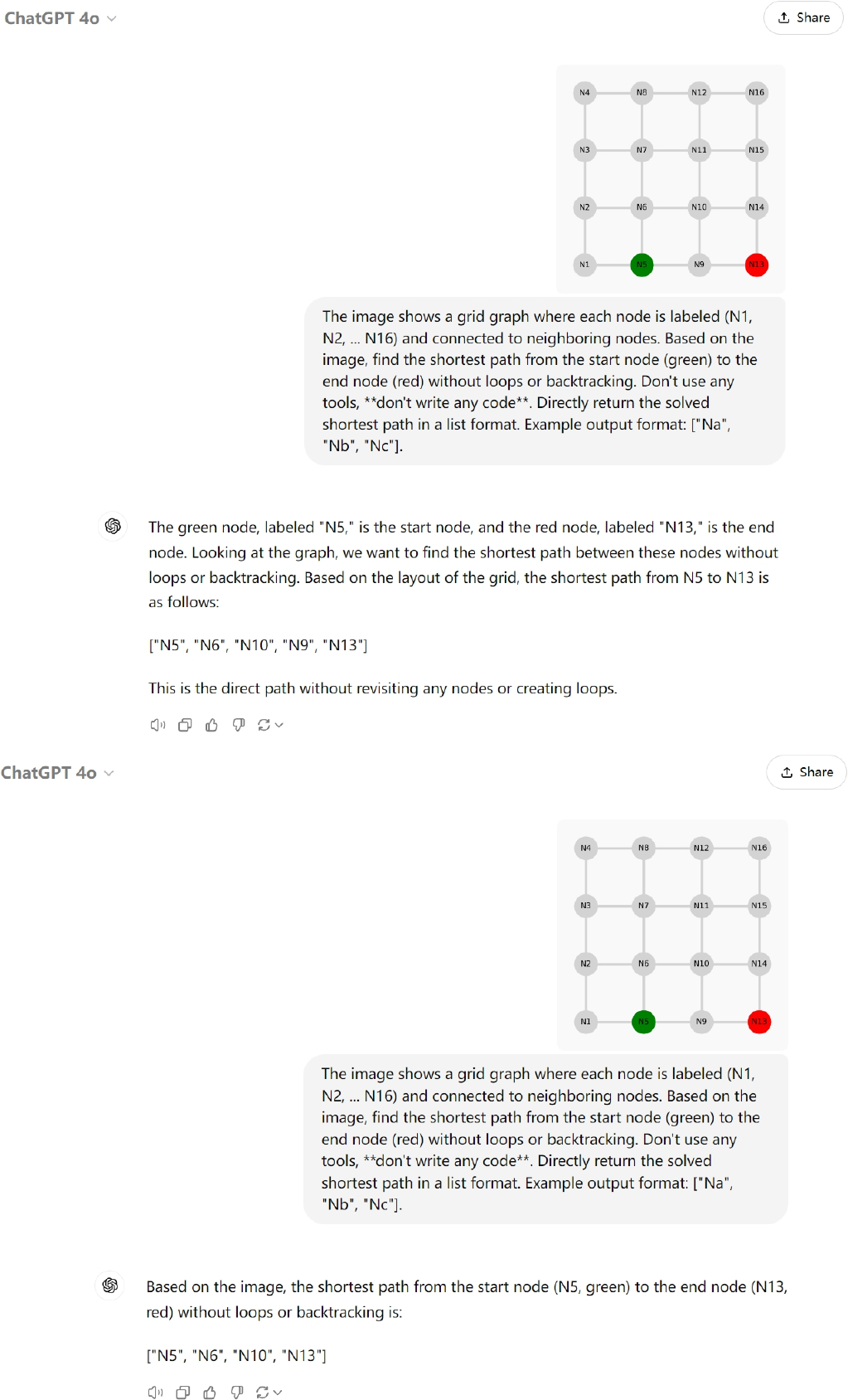}
    \caption{Screenshot supporting Figure 1 in the main paper: Chat interactions with GPT-4o.}
    \label{fig:gpt4o_1}
\end{figure}

\clearpage

\subsection{InternVL2-Pro}

\begin{figure}[h]
  \centering
    \includegraphics[width=0.7\textwidth]{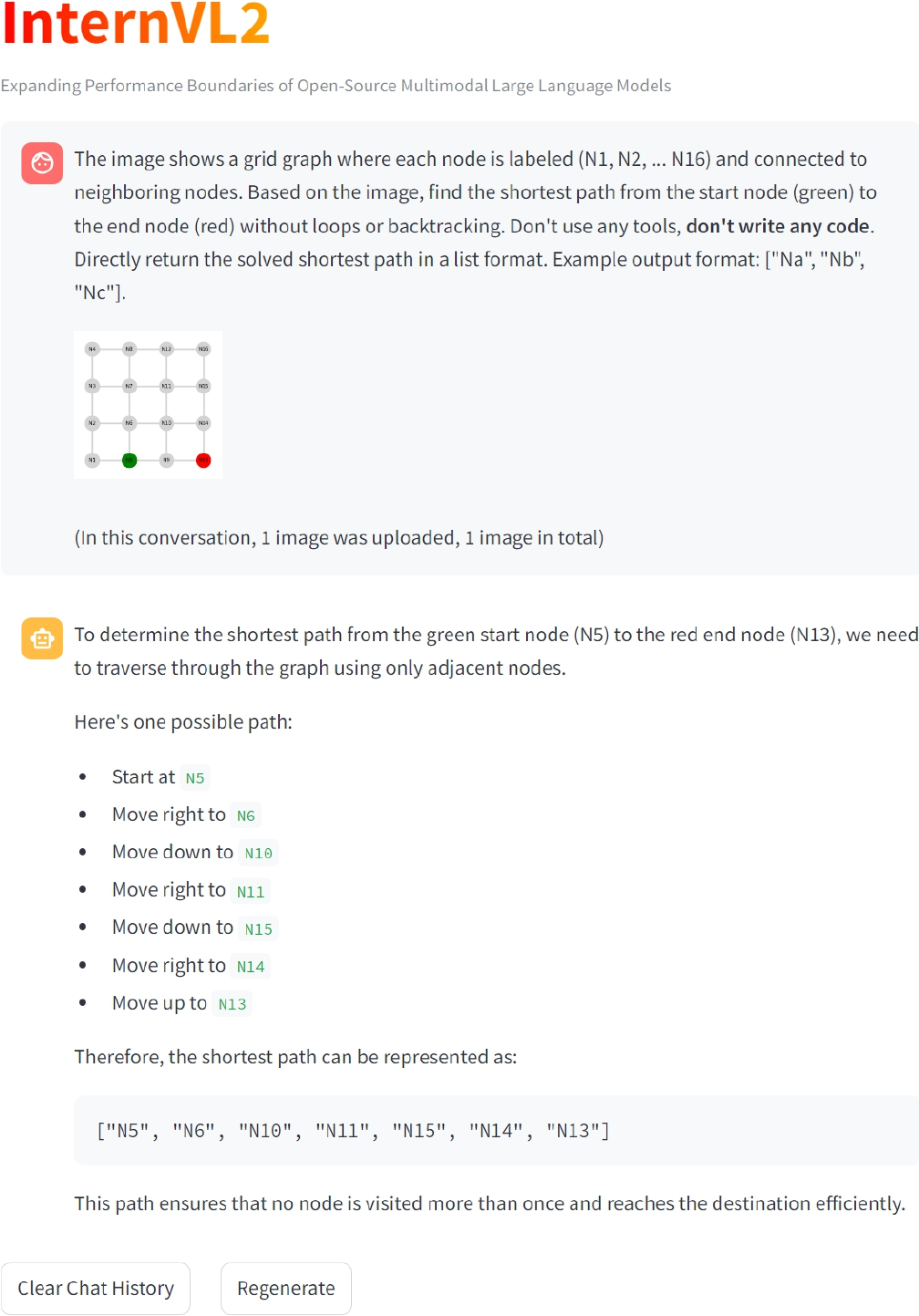}
    \caption{Screenshot supporting Figure 1 in the main paper: Chat interactions with InternVL2-Pro.}
    \label{fig:internvl2_pro}
\end{figure}

\end{onecolumn}

\end{document}